\documentclass[journal]{IEEEtran}
\usepackage{amsmath,graphicx, amssymb,bm}
\usepackage{subfigure}
\usepackage{graphicx}
\usepackage[ruled, linesnumbered]{algorithm2e}
\usepackage{wrapfig}
\usepackage{multirow}
\usepackage{cite}
\usepackage{url}
\usepackage{threeparttable}
\usepackage{booktabs} %
\usepackage{xcolor}
\usepackage{amsfonts}
\usepackage[colorinlistoftodos]{todonotes}
\usepackage[colorlinks=true, allcolors=blue]{hyperref}

\PassOptionsToPackage{dvipsnames}{xcolor}
\usepackage{color}
\usepackage{tikz}
\usetikzlibrary{arrows,positioning,fit,petri,shapes,backgrounds,decorations.pathmorphing,calc,shapes.misc, arrows, decorations.markings}

\newcommand{\eg}{e.\,g.,\ }
\newcommand{\ie}{i.\,e.,\ }

\newcommand{\et}{{et al.}}

\newcommand{\mb}{\mathbf}
\newcommand{\mc}{\mathcal}

\begin{document}
%
\title{Difficulty Learning/Weakness Learning for Continuous Emotion Recognition}
\title{Difficulty Sensitive Evolving Learning for Continuous Emotion Recognition}
\title{Dynamic Difficulty Aware/Sensitive Learning for \\Continuous Emotion Recognition}
\title{Dynamic Difficulty Attention Modelling for Continuous Emotion Prediction}
\title{Dynamic Difficulty Awareness Training for Continuous Emotion Prediction}

\author{Zixing~Zhang, Jing Han, Eduardo Coutinho,
and~Bj{\"o}rn~Schuller 
\thanks{Z.~Zhang is with GLAM -- the Group on Language, Audio \& Music, Imperial College London (UK). E-mail: zixing.zhang@imperial.ac.uk.}
\thanks{J.~Han (corresponding author) is with the ZD.B Chair of Embedded Intelligence for Health Care and Wellbeing, University of Augsburg (Germany). E-mail: jing.han@informatik.uni-augsburg.de.}
\thanks{E.~Coutinho is with the Department of Music, University of Liverpool (UK), and the ZD.B Chair of Embedded Intelligence for Health Care and Wellbeing, University of Augsburg (Germany). E-mail: e.coutinho@liverpool.ac.uk.}
\thanks{B.~Schuller is with GLAM -- the Group on Language, Audio \& Music, Imperial College London (UK), and the ZD.B Chair of Embedded Intelligence for Health Care and Wellbeing, University of Augsburg (Germany). E-mail: bjoern.schuller@imperial.ac.uk.}
}

%

\maketitle

\begin{abstract}
Time-continuous emotion prediction has become an increasingly compelling task in machine learning. Considerable efforts have been made to advance the performance of these systems. Nonetheless, the main focus has been the development of more sophisticated models and the incorporation of different expressive modalities (\eg speech, face, and physiology). In this paper, motivated by the benefit of difficulty awareness in a human learning procedure, we propose a novel machine learning framework, namely, Dynamic Difficulty Awareness Training (DDAT), which sheds fresh light on the research -- directly exploiting the difficulties in learning to boost the machine learning process. The DDAT framework consists of two stages: information retrieval and information exploitation. In the first stage, we make use of the reconstruction error of input features or the annotation uncertainty to estimate the difficulty of learning specific information. The obtained difficulty level is then used in tandem with original features to update the model input in a second learning stage with the expectation that the model can learn to focus on high difficulty regions of the learning process. We perform extensive experiments on a benchmark database (RECOLA) to evaluate the effectiveness of the proposed framework. The experimental results show that our approach outperforms related baselines as well as other well-established time-continuous emotion prediction systems, which suggests that dynamically integrating the difficulty information for neural networks can help enhance the learning process.

\end{abstract}


\IEEEpeerreviewmaketitle

\section{Introduction}
\label{sec:introduciton} 
\textit{Time-continuous emotion prediction} systems have received widespread interest in the machine learning (ML) community over the past decade~\cite{Woellmer08-Abandoning,Gunes10-Automatic,Gunes11-Emotion}.
One of the main reasons for this interest is the fact that time-continuous emotion predictions can analyse subtle and complex affective states of humans over time and play a central role in smart conversational agents that aim to achieve a natural and intuitive interaction between humans and machines~\cite{Gunes10-Automatic,Picard97-Affective,Yang13-Quantitative,Zhao17-Continuous,Zhang18-Leveraging}.
Great efforts have been made in this field, and most of them can generally be classified into two strands. One strand mainly focuses on designing or implementing increasingly sophisticated and robust prediction models, such as long short-term memory (LSTM)-based recurrent neural networks (RNNs)~\cite{Woellmer08-Abandoning,Schuller11-AVEC}, convolutional neural networks (CNNs)~\cite{Mao14-Learning,Meng16-Time,Xia17-Multi,Zhang17-Speech,Li17-Multimodal}, and end-to-end learning frameworks~\cite{Tzirakis17-End}. Another strand mainly focuses on the integration of multiple modalities (\eg, audio and video) and modelling techniques~\cite{Zeng09-Survey,Han17-Strength}.

Apart from those studies, other research has recently found that emotional training data can be practically learnt in different degrees~\cite{Zhang14-Agreement,Gui17-Curriculum}. That is, some data can be  easily learnt given a specific model, whilst some data are relatively tough. In this light, some promising approaches have been proposed in machine learning to optimise the learning procedure. For example, the most conventional approach is associated with boosting~\cite{Liu14-Facial, Presti17-Boosting}, which dynamically updates the weights of those samples that are hard to recognise or are even falsely recognised. 
Additionally, a more recent and promising approach refers to curriculum learning, which was firstly introduced in~\cite{Bengio09-Curriculum}. Curriculum learning presents the data from easy to hard during the training process so that the model can better avoid being caught in local minima in the presence of non-convex training criteria. Curriculum learning has become even more popular with the advance of deep learning. For emotion prediction, a handful of related studies have been reported very recently~\cite{Gui17-Curriculum,Braun17-curriculum,Lotfian18-Curriculum}, which have shown the efficiency of curriculum learning. 

However, one of the major disadvantages of these approaches is their non-friendliness to sequence-based pattern recognition tasks, such as the one we are facing. 
That is, in the learning process, the samples, whether or not they were presented within a sequence, are considered individually and independently. The ignored context information, nevertheless, indeed plays a vital role in sequence-based pattern recognition~\cite{Graves12-Supervised}. 
To this end, we propose a novel learning framework, \textit{Dynamic Difficulty Awareness Training} (DDAT), for time-continuous emotion prediction in this article. 
In contrast to the previous approaches, such as the aforementioned boosting and curriculum learning, the proposed DDAT can be well integrated into conventional context-sensitive models (\eg LSTM-RNNs), enabling the models to ultimately exploit the context information. 
To the best of our knowledge, this is the first effort at exploring the difficulty information in sequence-based pattern recognition, such as the present case of time-continuous emotion prediction.

The underlying assumption of DDAT is that a model is able to deliver better performance if we explicitly let the model know the learning difficulty of the samples along with time. 
This assumption is in line with the finding that humans normally pay more attention to the tasks that are inherently difficult so as to perform better~\cite{Posner90-Attention,Washburn01-Attention}. 

To implement DDAT, we consider two strategies, \ie utilising the \textit{Reconstruction Error} (RE) of the input data or the \textit{perception uncertainty} (PU) level of emotions as dynamic indicators of the difficulty to drive the learning process. Then, we integrate the difficulty indicator with original data for further learning, such that it endows the models with a difficulty learning awareness. This process is also partially inspired by the awareness techniques proposed for robust speech recognition~\cite{Seltzer13-investigation,Karanasou14-Adaptation}, where the noise types are considered to be auxiliary information for acoustic modelling. 

In ML, RE normally serves as an objective function of an auto-encoder (AE) when extracting high-level representations. A well-designed AE is considered to reconstruct well the input from its learnt high-level representations~\cite{Vincent08-Extracting}. Recently, the RE has also been exploited for tasks, such as anomaly detection~\cite{Marchi15-novel,Yang15-Unsupervised} and classification~\cite{Petridis16-Prediction}. For anomaly detection, an AE is trained on normal samples beforehand to serve as a novel event detector. When feeding a new sample into the AE, the obtained RE compared with a predefined threshold decides whether it is abnormal~\cite{Marchi15-novel,Yang15-Unsupervised}. For classification, several class-specific AEs are pre-trained separately. When feeding an unknown sample into these AEs simultaneously, the values of the corresponding RE are then interpreted as indicators of class membership~\cite{Petridis16-Prediction}. 
Notably, all these works hypothesise that data with the same label have similar data distributions. That is, the mismatched data potentially result in higher REs than those of the matched data. This motivates us to employ the RE as a learning difficulty index because it is well known in ML that mismatched data severely promote the complexity of modelling~\cite{Zhang17-Efficient}. 

In addition, PU is a term employed in subjective pattern recognition tasks to refer to the inter-annotator disagreement level when calculating a gold standard in an annotation process~\cite{Devillers05-Challenges}. 
For emotion prediction, it has been frequently determined that the PU has a high correlation with the learning difficulty of a recognition model. For example, the reported work in~\cite{Deng12-Confidence} and~\cite{Dang17-Investigation} found that the emotion prediction systems perform  better in low-uncertainty regions than in high-uncertainty regions. Likewise, the findings in~\cite{Zhang14-Agreement} showed that the elimination of the samples labelled with a high uncertainty from the training set leads to a better emotion prediction model. This finding provokes us to use the PU as another learning difficulty index. It is also worth noting that the principle of PU-based strategy constrains its application to subjective pattern recognition tasks. Despite the fact that the concept of `uncertainty' was employed in previous emotion prediction work, it was calculated among multiple predictions from variable systems~\cite{Dang17-Investigating}, which significantly differs from the definition of PU in this article, or merely utilised for multi-task learning~\cite{Han17-From} (cf.~Section~\ref{sec:relatedWork}). 

Motivated by the above analysis and following our previous tentative work~\cite{Han17-Reconstruction}, where only the RE was investigated for emotion prediction in speech, we demonstrate in this paper that the proposed DDAT framework can aid the ML models in detection of `moments' in the learning process that are of higher difficulty in the context of audiovisual time-continuous emotion prediction.
More specifically, the contributions of the present article include the following:
(i) proposing a new framework that exploits knowledge about the learning difficulty of the samples during the learning process for time-continuous emotion prediction;
(ii) introducing and analysing two specific strategies (\ie based on RE or PU) to implement this framework;
(iii) presenting a dynamic tuning approach to further dynamically tune the predictions; and
(iv) comprehensively evaluating the effectiveness of the proposed framework on a benchmarked audiovisual emotion prediction database.

The remainder of this article is organised as follows.
In Section~\ref{sec:relatedWork}, we briefly review past and related studies.
In Section~\ref{sec:methodology}, we present a detailed description of the structure and algorithm of the proposed DDAT framework.
Then, in Section~\ref{sec:experimentsResults}, we offer an extensive set of experiments conducted to exemplify the effectiveness and robustness of the DDAT framework along with a discussion.
Finally, we present our conclusions and future research directions in Section~\ref{sec:conclusion}.

\section{Related Work}
\label{sec:relatedWork}
For continuous emotion prediction, plenty of novel approaches have been proposed and investigated over the past decade.   
Some approaches expect to design or implement a more sophisticated and robust prediction model~\cite{Schuller11-AVEC,Woellmer08-Abandoning,Mao14-Learning,Meng16-Time,Xia17-Multi,Tzirakis17-End}.
Given that context information is crucial for estimating sequential patterns (continuous emotion prediction in our case), recurrent neural networks (RNNs), especially the ones implemented with long short-term memory (LSTM) cells, were introduced~\cite{Woellmer08-Abandoning}, and they are still amongst current state-of-the-art models~\cite{Brady16-Multi}. 
One of the main advantages of LSTM-RNNs is that they can model long-range dependencies between sequences ~\cite{Hochreiter97-Long,Graves12-Supervised}, and therefore, they are efficient in capturing the temporal information of emotional expression~\cite{Woellmer08-Abandoning}. More recently, the so-called end-to-end network architecture has been emerging as a promising network structure, which can automatically derive representations directly from raw (unprocessed) data, rather than manually extracting hand-crafted features. For example, in~\cite{Tzirakis17-End}, Tzirakis~\et~jointly trained the CNNs at the front end and the LSTM-RNNs at the back end, where the CNNs mainly take charge of extracting representations from raw audio signals and the concatenated LSTM-RNNs are responsible for capturing the temporal information. A similar framework has also been shown in~\cite{Ma16-DepAudioNet:}. 

Meanwhile, some other approaches attempt to overcome the drawbacks of individual models by means of integrating multiple different modalities or models in an ensemble strategy~\cite{Zeng09-Survey,Han17-Strength}. 
One common approach when considering multiple modalities is \textit{early} (aka \textit{feature}-level) fusion of unimodal information.
This is typically achieved by concatenating all the features from multiple modalities into one combined feature vector, which is then used as the input information for the models~\cite{Ringeval15-Prediction,Han17-Strength, Chao15-Long, Huang15-Investigation}.
A benefit of early fusion is that it can provide better discriminative ability to the model by exploiting the complementary information that exists among different modalities.
For example, acoustic features empirically outperform visual features for arousal estimation, whereas the opposite occurs for valence estimation~\cite{Ringeval15-Prediction}.
Another frequently employed approach is \textit{late} (aka \textit{decision}-level) fusion, which involves the combination of predictions obtained from diverse learners (models) to determine the final prediction.
To build the diverse learners,  Wei~\et~\cite{Wei14-Multimodal} created an ensemble of LSTM-RNN learners that were trained on different modalities (\eg audio and video), whereas Qiu~\et~\cite{Qiu14-Ensemble} developed a variety of topology structures of deep belief networks (DBN). 
To combine the predictions from multiple learners, a straightforward approach applies (un-)weighted averaging, such as simple linear regression (SLR)~\cite{Valstar16-AVEC, Huang15-Investigation}.
Another common approach is stacking, whereby the predictions from different learners are stacked and used as inputs of a subsequent non-linear model that is trained to make a final decision~\cite{He15-Multimodal,Wei14-Multimodal, Qiu14-Ensemble}.
In order to leverage the individual advantages of different models, Han~\et~\cite{Han17-Strength} further proposed a \textit{strength modelling} framework that concatenates two different models in a hierarchical architecture.
In this approach, the prediction yielded by the first model is concatenated with the original input features, and this expanded feature vector is then set as the input to the next model.

All of the outlined approaches above merely focus on either extending the capability or overcoming the drawbacks of the learning model. Difficulty information in the learning process, however, has seldom been exploited to date, to the best of our knowledge. 

Moreover, DDAT relates to \textit{multi-task learning} (MTL) as well~\cite{Nicolaou11-Continuous,Deng18-Semi,Han17-From,Zhao17-Continuous}.
In~\cite{Deng18-Semi}, Deng~\et~reconstructed the inputs with an AE as an auxiliary task for emotion prediction in a semi-supervised manner, and they demonstrated that the AE can distill representative high-level features from large-scale unlabelled data.
In~\cite{Han17-From}, Han~\et~proposed utilisation of the PU as an auxiliary task for continuous and dimensional emotion prediction, and they found that this information helps improve performance. 
In~\cite{Nicolaou11-Continuous}, Nicolaou~\et~introduced an output-associative framework to learn the correlations and patterns among different emotional dimensions (\ie arousal \textit{and} valence).
In this framework, the arousal and valence predictions from independent models are fused together and fed into a consequential model for a final prediction (\ie arousal \textit{or} valence).
The effectiveness of this approach has been replicated in~\cite{Nicolaou12-Output, Parthasarathy17-Jointly}.

Analogous to MTL, the present DDAT framework considers the tasks of reconstructing inputs or predicting perception uncertainty to be auxiliary tasks. Nevertheless, the RE and the PU are further assumed to be the learning difficulty indicators, and the model inputs are dynamically updated in order to endow the model with a difficulty-aware learning capability.

\section{Dynamic Difficulty Awareness Training}
\label{sec:methodology}

In this section, we describe the DDAT framework. Let $\mb{x}\in\mathcal{X}$ denote the feature vector in the input feature space, and $y\in\mathcal{Y}$, the label in the emotion label space. For a sequential pattern recognition task in our case, $\mb{x}_t$ thus indicates a feature vector at the $t$-th frame extracted from an utterance.

\subsection{System Overview}

The pseudo-code describing the proposed algorithm is presented in Algorithm~\ref{alg:reDFL}. It consists of two main stages: (i) \textit{retrieving} difficulty information and (ii) \textit{exploiting} difficulty information. In the first stage, in order to extract and indicate the information related to the difficulty of the learning process, we propose two different strategies: \textit{ontology-} and \textit{content-driven} strategies.

The \textit{ontology-driven} strategy focuses on the model itself. 
Specifically, we determine the difficulty of the task through the reconstruction of the input information, assuming that the RE is a proxy for its learning capability in a given moment. 

On the contrary, the \textit{content-driven} strategy focuses on the data and assumes that different data can be learnt to different degrees. That is, some data can be easily learnt with a specific model, whereas other data can be difficult. This approach partially stems from curriculum learning~\cite{Bengio09-Curriculum}, which has demonstrated that each datum cannot be equally learnt so as to be well-organised for model training.
In the field of emotion prediction, a few studies have shown that the difficulty-level of the data to be learnt is closely related to its PU~\cite{Deng12-Confidence,Dang17-Investigation}, as discussed in Section~\ref{sec:introduciton}.
Inspired by these studies, we employ the PU to represent the difficulty and complexity of the samples.

In the second stage, we concatenate the original features $\mb{x}_t$ with the difficulty vector $\mb{d}_t$ retrieved by one of the aforementioned two strategies, update the inputs via $[\mb{x}_t, \mb{d}_t]$ and re-train the regression model for continuous emotion prediction.
Due to the fact that $\mb{d}_t$ varies along with time, the extended difficulty vector provides dynamic awareness when modelling $\mb{x}$ in a continuum.

  \begin{algorithm}[!t]

  \SetKwInput{Initialize}{Initialise}
  \Initialize{\\
    $h$: neural networks; \\
    $\mb{x}$: feature vector, $\mb{x} = [x_1, x_2, \ldots, x_r]$ 
    $I, J$: predefined training epochs
    }

  \vspace{0.3cm}
  \uIf{ontology-driven}{auxiliary task $\mathcal{T} \leftarrow$ reconstructing input; }
  \uElseIf{content-driven}{auxiliary task $\mathcal{T} \leftarrow$ predicting perception uncertainty;}
  \textbf{end}

  \vspace{0.3cm}
  {\em\% retrieving difficulty information stage} \\

  \For{$i=1,...,I$}{

  optimise $h$ by minimising a joint loss function $\mc{J}(\bm{\theta}_0) = w_1*L_{emt}(\cdot) + w_2*L_{aux}(\cdot) + \lambda R(\bm{\theta}_0)$, 
  where $w_1$ and $w_2$ regulate the contributions of the emotion prediction $L_{emt}(\cdot)$ and the auxiliary task prediction $L_{aux}(\cdot)$; \\

  evaluate $h$ on the development set for emotion prediction: $CCC_{val, i}$; \\

  \If{$CCC_{val, i} > CCC_{val, i-1}$}{save $h$;}

  }

  obtain the difficulty attention $\mb{d}$ based on the chosen auxiliary task $\mathcal{T}$; 
  \vspace{0.3cm}

{\em\% exploiting difficulty information stage} \\
  \For{$j=1,...,J$}{
  update the input feature vector: $\mb{x}' = [\mb{x},\mb{d}]$; \\

  optimise $h$ by minimising the loss function $\mc{J}(\theta_{emt})$ for emotion prediction;

  evaluate $h$ on the development set for emotion prediction: $CCC_{val, j}$; \\

  \If{$CCC_{val, j} > CCC_{val, j-1}$}{save $h$;}

  }

    \caption{Dynamic Difficulty Awareness Training}
    \label{alg:reDFL}
  \end{algorithm}

\subsection{Multi-Task Learning}
\label{subsec:MTL}

MTL is a process of learning multiple tasks \textit{concurrently}.
Typically, there is one main task and one or more auxiliary tasks.
By attempting to model the auxiliary tasks together with the main task, the model learns shared information among tasks, which may be beneficial to learning the main task.
Mathematically, the objective function in MTL can be formatted as:
\begin{equation}
 \mc{J}(\bm\theta_0)={\sum_{m=1}^{M}w_m L_m(\mb{x},{y}_{m};\bm{\theta}_m)+{\lambda}R(\bm{\theta}_0)},
\label{eq:tl1}
\end{equation}
where $M$ denotes the number of tasks and $L_m(\cdot)$ represents the loss function of the task $m$, which is weighted by $w_m$.
$\bm{\theta}_0$ and $\bm{\theta}_m$ represent, respectively, the general model parameters and the specific ones with respect to task $m$, and $\lambda$ is a hyperparameter that controls the importance of the regularisation term $R(\bm{\theta}_0)$.

In this article, in order to infer the difficulty of the information being modelled in the first stage of the DDAT framework, we use an MTL structure to jointly learn continuous emotion prediction together with the reconstruction of the input features or the PU prediction.
The rationale is twofold:
On the one hand, the model makes better use of MTL for continuous emotion prediction.
The benefit of MTL has been shown by several studies for emotion prediction, as described in Section~\ref{sec:relatedWork}. 
On the other hand, the model takes one network, rather than two~\cite{Han17-Reconstruction}, to explore the difficulty of the learning process.

\subsection{Ontology-Based Difficulty Information Retrieval} 
\label{subsec:re}

  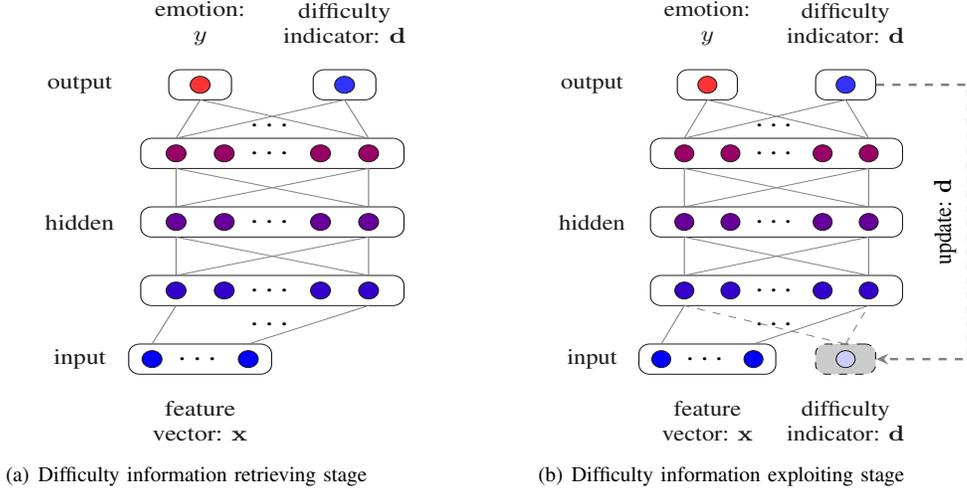
\begin{figure*}[!t]
  \begin{center}
   \hspace{-1.5cm}
    \subfigure[Difficulty information retrieving stage]{
    \resizebox{0.75\columnwidth}{2.4in}{
    \large \
    %
%
%
%
%
%
%
%
 
\tikzstyle{neuron}=[circle,draw=black!90, fill=#1, minimum size=7pt,inner sep=0pt]
\tikzstyle{object} = [rectangle, draw=black!90,  text width=3cm, minimum height=.44cm, rounded corners, font=\small]
\tikzstyle{connect} = [color=black!90, line width=0.8pt]
\tikzstyle{symbol} = [color=black!90, align = center, node distance=1.5cm, font=\small]
 
\def\layersep{0.5cm}
\def\neuronsep{0.6cm}
\begin{tikzpicture}[>=stealth,node distance = 1.5cm,auto]

       \foreach \name / \y in {20,40,60}
          \node [object] (b-\name) at (2*\neuronsep,\layersep*\y/10) {}; 
       \foreach \name / \y in {0}
          \node [object, text width=1.5cm] (bi-1) at (0.5*\neuronsep,\layersep*\y/10) {}; 
       \foreach \name / \y in {80}
          \node [object, text width=0.5cm] (bo-1) at (0.5*\neuronsep,\layersep*\y/10) {}; 
       \foreach \name / \y in {80}
          \node [object, text width=0.5cm] (bo-2) at (3.5*\neuronsep,\layersep*\y/10) {};

       \foreach \name / \y in {20,40,60}
	  \node[neuron={red!\y!blue}] (n1-\name) at (0,\layersep*\y/10) {};
       \foreach \name / \y in {20,40,60}
	  \node[neuron={red!\y!blue}] (n2-\name) at (1*\neuronsep,\layersep*\y/10) {};
       \foreach \name / \y in {20,40,60}
          \node[](n3-\name) at (2*\neuronsep,\layersep*\y/10) {$\cdots$}; 
       \foreach \name / \y in {20,40,60}
	  \node[neuron={red!\y!blue}] (n4-\name) at (3*\neuronsep,\layersep*\y/10) {};
       \foreach \name / \y in {20,40,60}
	  \node[neuron={red!\y!blue}] (n5-\name) at (4*\neuronsep,\layersep*\y/10) {};
	
       \foreach \name / \x in {0, 2} 
          \node[neuron={blue}] (ni-\name) at ({\x*\neuronsep-0.5*\neuronsep}, 0) {} ; 
       \foreach \name / \x in {1} 
          \node[] (ni-\name) at ({\x*\neuronsep-0.5*\neuronsep}, 0) {$\cdots$} ; 
	  
       \foreach \name / \x in {1} 
          \node[neuron={red!80!white}] (no-\name) at ({\x*\neuronsep-0.5*\neuronsep},  \layersep*8) {} ; 
       \foreach \name / \x in {4} 
          \node[neuron={blue!80!white}] (no-\name) at ({\x*\neuronsep-0.5*\neuronsep},  \layersep*8) {} ; 

       \foreach \name / \y in {10, 68}
	  \node [] () at ((2*\neuronsep,\layersep*\y/10){$\cdots$};

       \draw [-, solid, draw=gray] ($(n1-20.north)+(0,0.9mm)$) to ($(n1-40.south)+(0,-0.9mm)$);
       \draw [-, solid, draw=gray] ($(n1-20.north)+(0,0.9mm)$) to ($(n5-40.south)+(0,-0.9mm)$);
       \draw [-, solid, draw=gray] ($(n5-20.north)+(0,0.9mm)$) to ($(n1-40.south)+(0,-0.9mm)$);
       \draw [-, solid, draw=gray] ($(n5-20.north)+(0,0.9mm)$) to ($(n5-40.south)+(0,-0.9mm)$);
       
       \draw [-, solid, draw=gray] ($(n1-40.north)+(0,0.9mm)$) to ($(n1-60.south)+(0,-0.9mm)$);
       \draw [-, solid, draw=gray] ($(n1-40.north)+(0,0.9mm)$) to ($(n5-60.south)+(0,-0.9mm)$);
       \draw [-, solid, draw=gray] ($(n5-40.north)+(0,0.9mm)$) to ($(n1-60.south)+(0,-0.9mm)$);
       \draw [-, solid, draw=gray] ($(n5-40.north)+(0,0.9mm)$) to ($(n5-60.south)+(0,-0.9mm)$);
       
       \draw [-, solid, draw=gray] ($(ni-0.north)+(0,0.9mm)$) to ($(n1-20.south)+(0,-0.9mm)$);
       \draw [-, solid, draw=gray] ($(ni-2.north)+(0,0.9mm)$) to ($(n5-20.south)+(0,-0.9mm)$);
       
       
       \draw [-, solid, draw=gray] ($(n1-60.north)+(0,0.9mm)$) to ($(no-1.south)+(0,-0.9mm)$);
       \draw [-, solid, draw=gray] ($(n5-60.north)+(0,0.9mm)$) to ($(no-1.south)+(0,-0.9mm)$);
       \draw [-, solid, draw=gray] ($(n1-60.north)+(0,0.9mm)$) to ($(no-4.south)+(0,-0.9mm)$);
       \draw [-, solid, draw=gray] ($(n5-60.north)+(0,0.9mm)$) to ($(no-4.south)+(0,-0.9mm)$);

       \node[symbol, below of=bi-1, node distance = 1.8*\layersep] (x) {feature\\vector: $\mb{x}$};
       \node[symbol, above of=bo-1, node distance = 1.8*\layersep] (y) {emotion:\\ $y$};
       \node[symbol, above of=bo-2, node distance = 1.8*\layersep] (xx) {difficulty\\indicator: $\mb{d}$};

       \node[symbol] (t-0) at ({0-2*\neuronsep}, 0) {input}; 
       \node[symbol] (t-20) at ({0-2*\neuronsep}, \layersep*40/10) {hidden}; 
       \node[symbol] (t-80) at ({0-2*\neuronsep}, \layersep*80/10) {output};

      
\end{tikzpicture}
 
    }}
    \subfigure[Difficulty information exploiting stage]{
    \resizebox{0.8\columnwidth}{2.4in}{
    \large \
    %
%
%
%
%
%
%
%
 
\tikzstyle{neuron}=[circle,draw=black!90, fill=#1, minimum size=7pt,inner sep=0pt]
\tikzstyle{object} = [rectangle, draw=black!90,  text width=3cm, minimum height=.44cm, rounded corners, font=\small]
\tikzstyle{connect} = [color=black!90, line width=0.8pt]
\tikzstyle{symbol} = [color=black!90, align = center, node distance=1.5cm, font=\small]
 
\def\layersep{0.5cm}
\def\neuronsep{0.6cm}
\begin{tikzpicture}[>=stealth,node distance = 1.5cm,auto]

       \foreach \name / \y in {20,40,60}
          \node [object] (b-\name) at (2*\neuronsep,\layersep*\y/10) {}; 
       \foreach \name / \y in {0}
          \node [object, text width=1.5cm] (bi-1) at (0.5*\neuronsep,\layersep*\y/10) {}; 
       \foreach \name / \y in {0}
          \node [object, text width=0.5cm, fill=black!20!white, dashed] (bi-2) at (3.5*\neuronsep,\layersep*\y/10) {}; 
       \foreach \name / \y in {80}
          \node [object, text width=0.5cm] (bo-1) at (0.5*\neuronsep,\layersep*\y/10) {}; 
       \foreach \name / \y in {80}
          \node [object, text width=0.5cm] (bo-2) at (3.5*\neuronsep,\layersep*\y/10) {};

       \foreach \name / \y in {20,40,60}
	  \node[neuron={red!\y!blue}] (n1-\name) at (0,\layersep*\y/10) {};
       \foreach \name / \y in {20,40,60}
	  \node[neuron={red!\y!blue}] (n2-\name) at (1*\neuronsep,\layersep*\y/10) {};
       \foreach \name / \y in {20,40,60}
          \node[](n3-\name) at (2*\neuronsep,\layersep*\y/10) {$\cdots$}; 
       \foreach \name / \y in {20,40,60}
	  \node[neuron={red!\y!blue}] (n4-\name) at (3*\neuronsep,\layersep*\y/10) {};
       \foreach \name / \y in {20,40,60}
	  \node[neuron={red!\y!blue}] (n5-\name) at (4*\neuronsep,\layersep*\y/10) {};
	
       \foreach \name / \x in {0, 2} 
          \node[neuron={blue}] (ni-\name) at ({\x*\neuronsep-0.5*\neuronsep}, 0) {} ; 
       \foreach \name / \x in {4} 
          \node[neuron={white!80!blue}] (ni-\name) at ({\x*\neuronsep-0.5*\neuronsep}, 0) {} ; 
       \foreach \name / \x in {1} 
          \node[] (ni-\name) at ({\x*\neuronsep-0.5*\neuronsep}, 0) {$\cdots$} ; 
	  
       \foreach \name / \x in {1} 
          \node[neuron={red!80!white}] (no-\name) at ({\x*\neuronsep-0.5*\neuronsep},  \layersep*8) {} ; 
       \foreach \name / \x in {4} 
          \node[neuron={blue!80!white}] (no-\name) at ({\x*\neuronsep-0.5*\neuronsep},  \layersep*8) {} ; 

       \foreach \name / \y in {10, 68}
	  \node [] () at ((2*\neuronsep,\layersep*\y/10){$\cdots$};

       \draw [-, solid, draw=gray] ($(n1-20.north)+(0,0.9mm)$) to ($(n1-40.south)+(0,-0.9mm)$);
       \draw [-, solid, draw=gray] ($(n1-20.north)+(0,0.9mm)$) to ($(n5-40.south)+(0,-0.9mm)$);
       \draw [-, solid, draw=gray] ($(n5-20.north)+(0,0.9mm)$) to ($(n1-40.south)+(0,-0.9mm)$);
       \draw [-, solid, draw=gray] ($(n5-20.north)+(0,0.9mm)$) to ($(n5-40.south)+(0,-0.9mm)$);
       
       \draw [-, solid, draw=gray] ($(n1-40.north)+(0,0.9mm)$) to ($(n1-60.south)+(0,-0.9mm)$);
       \draw [-, solid, draw=gray] ($(n1-40.north)+(0,0.9mm)$) to ($(n5-60.south)+(0,-0.9mm)$);
       \draw [-, solid, draw=gray] ($(n5-40.north)+(0,0.9mm)$) to ($(n1-60.south)+(0,-0.9mm)$);
       \draw [-, solid, draw=gray] ($(n5-40.north)+(0,0.9mm)$) to ($(n5-60.south)+(0,-0.9mm)$);
       
       \draw [-, solid, draw=gray] ($(ni-0.north)+(0,0.9mm)$) to ($(n1-20.south)+(0,-0.9mm)$);
       \draw [-, solid, draw=gray] ($(ni-2.north)+(0,0.9mm)$) to ($(n5-20.south)+(0,-0.9mm)$);
       
       \draw [-, dashed, draw=gray] ($(ni-4.north)+(0,0.9mm)$) to ($(n1-20.south)+(0,-0.9mm)$);
       \draw [-, dashed, draw=gray] ($(ni-4.north)+(0,0.9mm)$) to ($(n5-20.south)+(0,-0.9mm)$);
       
       \draw [-, solid, draw=gray] ($(n1-60.north)+(0,0.9mm)$) to ($(no-1.south)+(0,-0.9mm)$);
       \draw [-, solid, draw=gray] ($(n5-60.north)+(0,0.9mm)$) to ($(no-1.south)+(0,-0.9mm)$);
       \draw [-, solid, draw=gray] ($(n1-60.north)+(0,0.9mm)$) to ($(no-4.south)+(0,-0.9mm)$);
       \draw [-, solid, draw=gray] ($(n5-60.north)+(0,0.9mm)$) to ($(no-4.south)+(0,-0.9mm)$);

       \node[symbol, below of=bi-1, node distance = 1.8*\layersep] (x) {feature\\vector: $\mb{x}$};
       \node[symbol, below of=bi-2, node distance = 1.8*\layersep] (e) {difficulty\\indicator: $\mb{d}$};
       \node[symbol, above of=bo-1, node distance = 1.8*\layersep] (y) {emotion:\\ $y$};
       \node[symbol, above of=bo-2, node distance = 1.8*\layersep] (xx) {difficulty\\indicator: $\mb{d}$};

       \node[symbol] (t-0) at ({0-2*\neuronsep}, 0) {input}; 
       \node[symbol] (t-20) at ({0-2*\neuronsep}, \layersep*40/10) {hidden}; 
       \node[symbol] (t-80) at ({0-2*\neuronsep}, \layersep*80/10) {output};

       \draw [->, line width=1pt, dashed, gray] (bo-2.east) -- ($(bo-2.east)+(12mm,0)$)  -- node[left=5mm, above=0mm, rotate=90, black, font=\small]{update: $\mb{d}$}($(bi-2.east)+(12mm,0)$) -- ($(bi-2.east)$); 
      
\end{tikzpicture}
 
    }}
    \caption{Dynamic Difficulty Awareness Training (DDAT) includes difficulty information (a) retrieving stage and (b) exploiting stage.  Difficulty information is indicated by either the input reconstruction error (\ie an error vector or the sum of all errors), or the emotion perception uncertainties.}
    \label{fig:outline}
  \end{center}
  \end{figure*}

Figure~\ref{fig:outline} illustrates the framework of RE-based DDAT, where the difficulty indicator $\mb{d}$ is generated from the reconstruction process of the inputs. 
As described in Section~\ref{subsec:MTL}, the employed network is trained in an MTL context, and so the output includes two paths--the emotion prediction path and the AE path.
The former is trained in a supervised fashion, whereas the latter is trained in an unsupervised manner. Thus, there are two tasks to be conducted when training the network, \ie  predicting emotions and reconstructing inputs.
Specifically, given a time sequence as input $\mb{x}$, the network is optimised by minimising the loss function as
  \begin{equation}
    \begin{aligned}
     \mathcal{J}(\bm{\theta_0}) = w_1*L_{emt}(\cdot) + w_2*L_{re}(\cdot) + \lambda R(\bm{\theta_0}),
    \label{eq:re}
    \end{aligned}
  \end{equation}
where $L_{emt}(\cdot)$ and $L_{re}(\cdot)$ denote the loss functions for emotion prediction and input reconstruction, respectively. To calculate them, we take the mean square error (MSE) for both learning paths, \ie
for emotion prediction,
\begin{equation}
 L_{emt}(\cdot) = \sum_{t=1}^{T}||\hat{y}_t - y_t||^2;
\end{equation}
and for the input reconstruction,
\begin{equation}
 L_{re}(\cdot) = \sum_{t=1}^{T}||\hat{\mb{x}}_t - \mb{x}_t||^2,
\end{equation}
where ${\mb{x}_t}$ and $y_t$ are a sample and its annotation at time $t$ from an input sequence with a period of time $T$, respectively. $\hat{\mb{x}}_t$ and $\hat{y}_t$ denote the network predictions to reconstruct its inputs $\mb{x}_t$ and estimate the emotions $y_t$, respectively.

It is expected that $L_{re}(\cdot)\rightarrow 0$ if the model is sufficiently powerful and robust. 
However, empirical experiments have shown that the results are far from this expectation. Previous findings frequently indicate that a higher distribution mismatch between the given data and the entire training dataset is inclined to produce a higher RE~\cite{Xia15-Learning,Marchi15-novel,Yang15-Unsupervised,Zhang16-Facing}. Therefore, the RE somewhat implies the difficulty degree of the model to learn such data or, in other words, reflects the difficulty of the data to be learnt by the model.

Once the model is trained, the difficulty of the learning process ($\mb{d}$) can be obtained by computing the distance between the input $\mb{x}$ and its corresponding reconstruction $\hat{\mb{x}}$.
The distance can be either a vector $\mb{e}$ calculated by,
\begin{equation}
 \mb{d} = \mb{e} = \mb{x} - \hat{\mb{x}},
\end{equation}
or a scalar $E$ summed over all attributes, \ie
\begin{equation}
 \mb{d} = [E] = [\sum_{i=1}^r(x_i-\hat{x}_i)],
\end{equation}
where $\mb{x} = [x_1, x_2, \ldots, x_r]$ and $r$ is the dimension of the feature vector.

In the difficulty exploitation stage, we update the model input with the new vector, \ie $\mb{x}' = [\mb{x}, \mb{e}]$ or $\mb{x}' = [\mb{x}, E]$. In doing this, the input feature vectors are of $2r$ or $r+1$ dimensions when feeding back an error vector or scalar.

\subsection{Content-Based Difficulty Information Retrieval} 
\label{subsec:pu}

As mentioned earlier, \textit{PU} is an indicator of the uncertainty level of the perception of an emotional state for a given observed sample.
In the context of affective computing, we deem that emotion prediction is a subjective task that differs from many other objective pattern recognition tasks, such as face recognition, that hold a ground truth~\cite{Schuller13-Computational}.
In order to obtain a gold standard for a subjective task, it is required that a sufficient number of raters observe the same sample and that their ratings are collapsed in order to eliminate as much as possible individual variations in perception and rating.
In this case, a possible way to infer uncertainty is by calculating the \textit{inter-rater disagreement level}, which assumes that for each sample, the personal PU is highly correlated with the inter-rater disagreement level~\cite{Mauss09-Measures,Han17-From}.

In this study, the PU $u^{(i)}$, $i\in\{arousal, valence\}$, is represented by the standard deviation of the $K$ annotations as
  \begin{equation}
    u_n^{(i)}=\sqrt{\frac{1}{K-1}\sum_{k=1}^K(y_{n,k}^{(i)}-\bar{y}_{n}^{(i)})^2},
  \end{equation}
where $\bar{y}_n^{(i)}$ denotes the mean value given $K$ annotations:
  \begin{equation}
  \bar{y}_{n}^{(i)}=\frac{1}{K}\sum_{k=1}^{K}y_{n,k}^{(i)}.
  \end{equation}

The framework of PU-based DDAT is also illustrated in Fig.~\ref{fig:outline}, where the difficulty indicator $\mb{d}$ is determined by the perception uncertainty.
The designed network includes an emotion prediction path and a PU prediction path, both of which are jointly trained in a supervised manner.
Therefore, the objective function of Eq.~\eqref{eq:tl1} can be re-formulated as
  \begin{equation}
    \begin{aligned}
     \mathcal{J}(\bm{\theta_0}) = w_1\cdot L_{emt}(*) + w_2\cdot L_{pu}(*) + \lambda \cdot R(\bm{\theta_0}).
    \label{eq:re}
    \end{aligned}
  \end{equation}
$L_{pu}(*)$ stands for the loss functions for PU prediction, and it is expressed by
\begin{equation}
 L_{pu}(*) = \sum_{t=1}^{T}||\hat{u}_t - u_t||,
\end{equation}
where $u_t$ is a PU value for the sample at time $t$ from input sequences with time $T$. 

Once the network is optimised in the first learning stage, its input will then evolve to $\mb{x}'=[\mb{x},u]$ with $r+1$ dimensions in the second learning stage.

\subsection{Late Fusion and Dynamic Tuning}
\label{subsec:voting}
As discussed in Section~\ref{sec:relatedWork}, late fusion approaches have been frequently shown to be effective for continuous emotion prediction~\cite{Valstar16-AVEC, Zeng09-Survey,Han17-Strength} due to the fact that complementary information can be provided by the various modalities or models~\cite{Valstar16-AVEC, Zeng09-Survey,Han17-Strength}.
In this light, we conduct a late fusion to combine the emotion predictions from different modalities, learning models, or a combination thereof.
The late fusion is performed with an SLR approach:
  \begin{equation}
  y=\epsilon+\sum{\gamma_i\cdot{y_{i}}},
  \end{equation}
where $y_i$ denotes the original prediction with the modality (\ie audio or video) or model $i$ (\ie RE- or PU-based DDAT), $\epsilon$ and $\gamma_i$ are the parameters estimated on the development set, and $y$ is the fused prediction.

Despite the effectiveness of SLR, this conventional fusion approach simply assumes that the predictions $y_{i,t}$ in a continuum are considered to be equally important for each prediction stream $y_i$.
This means that the parameter of $\gamma_i$ remains a constant in time, given a set of $y_i$, and therefore, this approach ignores the changes of the reliability of the predictions along time.
To address this problem, we further propose a \textit{dynamic tuning} strategy according to the reliability of predictions in time.

Mathematically, we applied an additional SLR on the original prediction $y_{i,t}$ and the corresponding difficulty indicator $d_{i,t}$ at time $t$:
\begin{equation}
y'_{i,t}=\epsilon + \gamma_i \cdot y_{i,t} + \gamma_d \cdot d_{i,t},
\end{equation}
where $d_{i,t}$ is represented by $E_t$ for the RE-based DDAT systems or $u_t$ for the PU-based DDAT systems.
Intuitively, the prediction is dynamically tuned by the difficulty information.

\section{Experiments and Results}
\label{sec:experimentsResults}

To evaluate the effectiveness of the proposed methods, we conducted extensive experiments with the benchmark database of the AudioVisual Emotion Challenges (AVEC) from 2015~\cite{Ringeval15-AV+EC} and 2016~\cite{Valstar16-AVEC}.

\subsection{Databases and Features}
\label{subsec:databases}
The multimodal corpus REmote COLlaborative and Affective interactions (RECOLA)~\cite{Ringeval13-Introducing} (a standard database of the AVEC challenges for audiovisual time-continuous emotion prediction~\cite{Ringeval15-AV+EC,Valstar16-AVEC}) was selected for our experiments due to its widespread use in this area.
This database was created to study socio-affective behaviours from multimodal data in the context of remote collaborative tasks.
It includes audiovisual (and physiological) recordings of spontaneous and natural interactions from 27 French-speaking participants whilst solving a collaborative task conducted in dyads via video conferencing.
The corpus is comprised of audio, video, and peripheral physiology recordings that were obtained synchronously and continuously over time.

In order to ensure speaker-independence for ML experiments, the corpus was divided into three partitions--training, development (validation), and testing--with each partition containing nine collaborative sessions.
This division is balanced in terms of gender, age, and mother tongue of the participants. 
The corpus contains value- and time-continuous annotations of two affective dimensions--arousal and valence--that were obtained from six French-speaking raters (three female) for the first five minutes of each audiovisual recording.
The obtained labels were then resampled at a constant frame rate of 40\,ms and averaged over all raters to create a `gold standard' for each instance.
Inter-rater disagreements were also computed for all instances~\cite{Ringeval13-Introducing}.
For our experiments, we only made use of audio and video signals.

  The acoustic and visual features employed in our experiments are the same sets used to compute the AVEC 2015 and 2016 baselines for fair comparison with other methods.
  The acoustic features consist of the extended Geneva Minimalistic Acoustic Parameter Set (\textit{eGeMAPS}~\cite{Eyben16-Geneva}).
  Since the RECOLA database contains long time-continuous signals and annotations, two functionals (arithmetic mean and standard derivation) were applied over the sequential low-level descriptors (LLDs, \eg pitch, loudness, energy, Mel Frequency Cepstral Coefficients, jitter, and shimmer) over a fixed window of 8\,s with a 40\,ms step.
This resulted in a set of 88 acoustic features per segment.

  In relation to the visual features, we utilised both the \textit{appearance} and \textit{geometric} standard features of the AVEC challenges.
  The appearance features were computed by using local Gabor binary patterns from three orthogonal planes through splitting the video into spatio-temporal video volumes.
  A feature reduction was then performed by applying a principal component analysis from a low-rank (up to rank 500) approximation, leading to 84 features representing 98\% of the variance.
  To extract the geometric features, 49 facial landmarks were firstly extracted from each frame and then aligned with a mean shape from stable points (located on the eye corners and on the nose region).
  This resulted in 316 features per frame: \ie 196 features were obtained by computing the difference between the coordinates of the aligned landmarks and those from the mean shape and between the aligned landmark locations in the previous and the current frame, and 71 were obtained by calculating the Euclidean distances (L2-norm) and the angles (in radians) between the points in three different groups. An additional 49 features correspond to the Euclidean distance between the median of the stable landmarks and each aligned landmark in a video frame.
  
  Similar to the acoustic features, the arithmetic mean and the standard derivation were computed over the sequential visual features of each frame using a sliding window of 8\,s with a step size of 40\,ms.
  This process led to 168 appearance and 632 geometric visual features.

  For full details on the database and feature sets, please refer to~\cite{Ringeval15-AV+EC,Valstar16-AVEC}.
  Note that we obtained 67.5\,k extracted segments in total for each partition (training, development, or test).

\subsection{Experimental Setup and Evaluation Metrics}
\label{subsec:setups}

  The implemented DDAT framework in our experiments consists of a deep RNN (DRNN) equipped with gated recurrent units (GRUs)~\cite{Cho14-properties}.
  GRUs are an alternative to long short-term memory units, which can also capture the long-term dependencies in sequence-based tasks and mitigate the effects of the vanishing gradient problem~\cite{Cho14-properties}.
  Compared to LSTM units, GRUs have fewer parameters due to the fact that they do not have separate memory cells and output gates, which results in a faster training process and a less-training-data demand for achieving a good generalisation.
  Most importantly, many empirical evaluations~\cite{Jozefowicz15-empirical} have indicated that GRUs perform as competitively as LSTM units.
  
The DRNN structure was optimised in terms of the number of hidden layers and the number of GRUs per layer in the development phase. We applied a search grid that is comprised of \{1, 3, 5, 7, 9\} hidden layers and \{40, 80, 120\} hidden units per layer. For each learning strategy, we always choose the best performing network structure in order to alleviate the impact of the variation of network structures on the system performance.
The training of the DRNNs was conducted using the Adam optimisation algorithm~\cite{Kingma14-Adam} with an initial learning rate of 0.001. To facilitate the training process, we set the size of mini-batch to four. 
Additionally, an online standardisation was applied to the input data by using the means and variations of the training set.

Additionally, as suggested in~\cite{Valstar16-AVEC}, annotation delay compensation was employed to compensate for the temporal delay between the observable cues and the corresponding annotations reported by the annotators~\cite{Mariooryad15-Correcting}. 
We identified this delay to be 2.4\,s, according to a series of experimental evaluations in~\cite{Ringeval17-AVEC}, and shifted the gold standard back in time with respect to the features for all modalities and tasks in our experiments.

In order to evaluate the performance of the models, we took the official metric of the AVEC 2015 and 2016 challenges--the {\em Concordance Correlation Coefficient} (CCC)~\cite{Ringeval15-AV+EC}:
  \begin{equation}
  r_c = \frac{2r\sigma_x\sigma_y}{\sigma^2_x+\sigma^2_y+(\mu_x - \mu_y)^2},
  \label{eq5}
  \end{equation}
  where $r$ represents {\em Pearson's correlation coefficient} between two time series (\eg prediction and gold-standard), $\mu_x$ and $\mu_y$ denote the mean of each time series, and $\sigma^2_x$ and $\sigma^2_y$ stand for the corresponding variances. 
  Compared with PCC, the CCC considers not only the shape similarity between two series but also the value precision. This is especially relevant for estimating the performance of time-continuous emotion prediction models, as both the trends as well as absolute prediction values are relevant for describing the performance of a model. The CCC metric falls in the range of [-1, 1], where $+$1 represents perfect concordance, $-$1 total discordance, and 0 no concordance at all.

 To refine the obtained prediction, we further performed a chain of post-processing, including median filtering, centering, scaling, and time-shifting, as suggested in~\cite{Ringeval15-AV+EC,Valstar16-AVEC}.
  The filtering window size $W$ (ranging from 0.12\,s to 0.44\,s at a rate of 0.08\,s) and the time-shifting delay $D$ (ranging from 0.04\,s to 0.60\,s at a step of 0.04\,s) were optimised using a grid search method. All the post-processing parameters were optimised on the development set and then applied to the test set. Therefore, those post-processing parameters had various settings for different tasks.

To compare the proposed DDAT approach with other related and state-of-the-art approaches, we further conducted \emph{curriculum learning}, as introduced in Section~\ref{sec:introduciton}. We particularly selected the criterion of `disagreement between annotators' (\ie PU in this article) as an example because it is appropriate for the task at hand and also superior to other criteria~\cite{Lotfian18-Curriculum}. To retain the optimised setups, we continued using the deep neural networks (DNNs) equipped with two hidden layers (1\,024 nodes per layer) and split the whole training set into five parts based on the PU levels. Moreover, we implemented it with GRU-RNNs as well for a fair performance comparison between the curriculum learning and the proposed DDAT.

  Finally, to statistically compare the various experiments conducted with the AVEC challenges baselines, we carried out the \textit{Fisher r-to-z transformation}~\cite{Cohen13-Applied}.
   In detail, given two distributions $X$ and $Y$ [the pairs ($X_i$, $Y_i$) $\sim$ i.i.d.] that have a bivariate normal distribution with correlation, the Fisher transformation $z$ is approximately normally distributed with mean
 \begin{equation}
 m = {1 \over 2}\ln \left({{1+r } \over {1-r }}\right)=\operatorname {arctanh} (r),
 \end{equation}
  and standard error
  \begin{equation}
   \sigma =  {1 \over {\sqrt {N-3}}},
  \end{equation}
  where $N$ is the sample size and $r$ is the true correlation coefficient. 
  Theoretically, the Fisher transformation is exceptionally efficient for small sample sizes because the sampling distribution of the Pearson correlation is normally highly skewed. 
  
  After the Fisher transformation, a one-tailed test was performed to compare two distributions. A $p$-value lower than $.05$ indicates a significant difference. It is noted that $r$ is replaced with $r_c$ (CCC) due to the efficiency of $r_c$ in this article.

\subsection{Emotion prediction with Dynamic Difficulty Awareness Training}
\label{subsec:ddat}

  \begin{table*}[!th]
  \centering
  \caption{System performance (Concordance Correlation Coefficient; CCC) {\bf before} post-processing the model predictions for the conventional single-task learning (baseline) framework, the multi-task learning (MTL) framework, and the proposed Dynamic Difficulty Awareness Training (DDAT) framework using reconstruction error (RE, a vector or a scalar of sum) and perception uncertainty (PU) variants. These results pertain to the experiments conducted on the \textit{dev}elopment and \textit{test} partitions for both {\em aro}usal and {\em val}ence targets. Three feature sets (audio-eGeMAPS, video-appearance, and video-geometric) were employed to evaluate all approaches. The cases where DDAT has a statistical significance of performance improvement over MTL are marked by the ``$\star$'' symbol.}
  \label{tab:beforePP}
    \begin{threeparttable}
    \begin{tabular}{llllllllllllllll}
    \toprule
    \multirow{3}{*}{\quad CCC}	     && \multicolumn{4}{c}{audio-eGeMAPS} && \multicolumn{4}{c}{video-appearance} && \multicolumn{4}{c}{video-geometric}      \\
		\cmidrule{3-6} \cmidrule{8-11} \cmidrule{13-16}
		&& \multicolumn{2}{c}{aro}   & \multicolumn{2}{c}{val}              && \multicolumn{2}{c}{aro}             & \multicolumn{2}{c}{val}   && \multicolumn{2}{c}{aro}   & \multicolumn{2}{c}{val}   \\
		&& dev   & test             & dev             & test  && dev   & test  & dev   & test  && dev & test & dev & test \\
    \midrule
    \em{baseline}    && .743 & .617            & .460           & .380 && .501 & .416 & .481 & .391 &&    .407 & .256 & .598 & .403 \\[3pt]
    \multicolumn{15}{l}{\em{Multi-task learning (MTL)}} \\
    \quad RE-based    && .743 & .590            & .513           & .298 && .472 & .434 & .512 & .351 &&     .429 & .283 & .632 & .487 \\
    \quad PU-based       && .727 & .613            & .485           & .417 && .459 & .426 & .444 & .342 &&     .442 & .284 & .630 & .444 \\[3pt]
    \multicolumn{15}{l}{\em{Proposed Dynamic Difficulty Awareness Training (DDAT) }} \\
    \quad RE-based (vector)  && .745 & .605$^\star$            & .485           & .374$^\star$ && .473 & .429 & .482 & .326 &&    .509$^\star$ & .299$^\star$ & .627  & .464 \\
    \quad RE-based (sum)     && .783$^\star$ & .671$^\star$            & .495           & .410$^\star$ && .487$^\star$ & .464$^\star$ & .507 & .460$^\star$ &&     .478$^\star$ & .359$^\star$ & .633 & .467 \\
    \quad PU-based      && .769$^\star$ & .623$^\star$            & .493           & .397 && .478$^\star$ & .457$^\star$ & .476$^\star$ & .412$^\star$ &&     .450 & .336$^\star$ & .629 & .500$^\star$ \\
    \midrule 
    \multicolumn{15}{l}{\em{Other state of the art}} \\
    \quad DNNs~\cite{Lotfian18-Curriculum} && .216 & .362 & .003 & .004 && .265 & .173 & .294 & .174 && .017 & .011 & .193 & .128 \\
	\quad Curriculum learning (DNN)~\cite{Lotfian18-Curriculum} && .445 & .533 & .018 & .006 && .402 & .292 & .437 & .444 && .199 & .116 & .271 & .245 \\
    \quad Curriculum learning (GRU-RNN) && .711	& .600	& .494	& .335 && .406	& .366	& .547	& .479	&& .392	& .291	& .582	& .497 \\ 
    \bottomrule
    \end{tabular}
    \end{threeparttable}
  \end{table*}

  \begin{table*}[!th]
  \centering
  \caption{System performance (Concordance Correlation Coefficient; CCC) {\bf after} post-processing the model predictions for the conventional single-task learning (baseline) framework, the multi-task learning (MTL) framework, and the proposed Dynamic Difficulty Awareness Training (DDAT) framework using reconstruction error (RE, a vector or a scalar of sum) and perception uncertainty (PU) variants. These results pertain to the experiments conducted on the \textit{dev}elopment and \textit{test} partitions for both {\em aro}usal and {\em val}ence targets. Three feature sets (audio-eGeMAPS, video-appearance, and video-geometric) were employed to evaluate all approaches. The best results achieved on the test set are in bold. The cases where DDAT has a statistical significance of performance improvement over MTL are marked by the ``$\star$'' symbol.}
  \label{tab:afterPP}
    \begin{threeparttable}
    \begin{tabular}{l@{}lllllllllllllll}
    \toprule
    \multirow{3}{*}{\quad CCC}	      && \multicolumn{4}{c}{audio-eGeMAPS} && \multicolumn{4}{c}{video-appearance} && \multicolumn{4}{c}{video-geometric}      \\
		\cmidrule{3-6} \cmidrule{8-11} \cmidrule{13-16}
		&& \multicolumn{2}{c}{aro}   & \multicolumn{2}{c}{val}              && \multicolumn{2}{c}{aro}             & \multicolumn{2}{c}{val}   && \multicolumn{2}{c}{aro}   & \multicolumn{2}{c}{val}   \\
		&& dev   & test             & dev             & test  && dev   & test  & dev   & test  && dev & test & dev & test \\
    \midrule
    \em{baseline}                && .783 & .652 & .473 & .400 && .528  & .403 & .493 & .404 &&     .523 & .314 & .620 & .417 \\[3pt]

    \multicolumn{15}{l}{\em{Multi-task learning (MTL)}} \\
    \quad RE-based            && .788 & .629 & .519 & .331 && .512 & .425 & .529 & .366 &&     .502 & .324 & .632 & .488 \\
    \quad PU-based            && .803 & .654 & .506 & .416 && .502 & .406 & .468 & .418 &&     .508 & .327 & .643 & .452 \\[3pt]

	    \multicolumn{15}{l}{\em{Proposed Dynamic Difficulty Awareness Training (DDAT)}} \\
    \quad RE-based (vector) && .806$^\star$ & .676$^\star$ & .517 & .378$^\star$ && .533$^\star$  & .434 & .520 & .329 &&  .559$^\star$ & .355$^\star$ & .634  & .473 \\
    \quad RE-based (sum)     && .807$^\star$ & \bf{.694}$^\star$ & .508 & \bf{.422}$^\star$ && .539$^\star$  & .437$^\star$ & .528 & \bf{.457}$^\star$ && .544$^\star$ & \bf{.400}$^\star$ & .639 & .471 \\
    \quad PU-based       && .811 & .664$^\star$ & .498 & .407 && .518$^\star$  & \bf{.438}$^\star$ & .514$^\star$ & .431$^\star$ &&      .513 & .397$^\star$ & .632 & \bf{.501}$^\star$ \\[3pt]
    
    \midrule
    
    \multicolumn{15}{l}{\em{Other state of the art}} \\
    \quad DNNs~\cite{Lotfian18-Curriculum} && .573 & .517 & .129 & .044 && .387 & .220  & .306 & .206 && .312 & .296 & .362 & .216 \\
	\quad Curriculum learning (DNN)~\cite{Lotfian18-Curriculum} && .687 & .591 & .159 & .174 && .417 & .343 & .446 & .419 && .394 & .267 & .300   & .269 \\
    \quad Curriculum learning (GRU-RNN) && .754	& .611 & .501	& .357 &&	.491 &	.391 &	.557 &	.492 &&	.444 &	.336 &	.609 &	.500 \\
    \quad Strength modelling~\cite{Han17-Strength}                   && .755 & .666 & .476 & .364 && .350            & .196 & .592 & .464 && --               & --     & --     & --     \\
    \quad End-to-end~\cite{Tzirakis17-End} \tnote{a} &&  .786 & .715     & .428 & .369     && .371    & .435     & .637 & .620    && --               & --     & --     & --  \\
    \quad Feature selection + offset~\cite{He15-Multimodal} \tnote{b}    && .800 & --     & .398 & --     && .587            & --     & .441 & --     && .173           & --     & .441 & --     \\
	\quad SVR + offset~\cite{Valstar16-AVEC} \tnote{c}   && .796 & .648 & .455 & .375 && .483            & .343 & .474 & .486 && .379           & .272 & .612 & .507 \\
    \quad SC + CNN + LSTM~\cite{Brady16-Multi} \tnote{d} && .846 & --     & .450 & --     && .346            & --     & .511 & --     && --               & --     & --     & --  \\

    \bottomrule
    \end{tabular}

    \begin{tablenotes}
    \scriptsize
    \item Note: ``--'' indicates that the corresponding CCC is not provided.
    \item[a]  acoustic and visual features automatically extracted by deep neural network models
	\item[b]  AVEC '15 challenge winner method
    \item[c]  AVEC '16 baseline method
    \item[d]  AVEC '16 challenge winner method
    \end{tablenotes}

    \end{threeparttable}
  \end{table*}

  The performance of the evaluated systems \textit{before} and \textit{after} post-processing the predictions for both arousal and valence targets is presented in Tables~\ref{tab:beforePP} and~\ref{tab:afterPP}, respectively.
To investigate the proposed DDAT framework, we not only conducted the traditional single-task learning but also the MTL for comparison, with three different feature sets--one acoustic feature set (eGeMAPS) and two visual feature sets (appearance and geometric features), as described in Section~\ref{subsec:databases}. 
It is worthy to note that the network structure employed for each modality and learning approach was respectively optimised in the constrained parameter space, as mentioned in Section~\ref{subsec:setups}. Then, the best performing network structures were employed for performance comparison. Doing this largely alleviates the inconsistent impact on the system performance due to the variation of network structures. 
  From the comparison of both Tables~\ref{tab:beforePP} and~\ref{tab:afterPP}, it can be seen that the post-processing of the model predictions generally leads to better performance. 
  For instance, the best baselines for arousal and valence are respectively boosted from 0.617 to 0.652 CCC with acoustic features (eGeMAPS) and from 0.403 to 0.417 CCC with visual features (geometric). Similar observations can also be obtained in the MTL systems and the proposed DDAT systems; \eg for the MTL systems, the CCCs are increased from 0.613 to 0.654 with the eGeMAPS feature set for arousal and from 0.487 to 0.488 with the geometric feature set for valence. 
Given these results, we henceforth focus on analysing the experiments with the post-processing step (cf.~Table~\ref{tab:afterPP}).

For the baseline system, the obtained results are competitive to, or even better than, 
the benchmark of the emotion prediction subchallenge in the AVEC 2016~\cite{Valstar16-AVEC} over three information streams and two prediction tasks. These results support previous findings showing that GRUs can deliver competitive performance when compared to LSTM units \cite{Cho14-properties,Jozefowicz15-empirical}.

\begin{table}[!t]
  \centering
  \caption{Obtained PCCs between each other among the performance improvement ($\Delta_c$), the reconstruction error ($\epsilon$), and the perception uncertainty ($\mu$). }
  \label{tab:relationship}
    \begin{threeparttable}
    \begin{tabular}{lccccc}
    \toprule
    && \multicolumn{2}{c}{aro}   & \multicolumn{2}{c}{val} \\
    && dev   & test   & dev   & test   \\
    \midrule
    \multicolumn{5}{l}{PCC($\epsilon$, $\Delta_c$)} \\
    \quad audio-eGeMAPS    && .128 & .180 & .078 & .114 \\
    \quad video-appearance    && .139 & .371 & .304 & .263 \\
    \quad video-geometric    &&  .140 &.155 &.044 &.090 \\
    \midrule
    \multicolumn{5}{l}{PCC($\mu$, $\Delta_c$)} \\
    \quad audio-eGeMAPS    && .150 & .181 & .150 & .181 \\
    \quad video-appearance    && .205 & .173 & .383 & .440 \\
    \quad video-geometric     &&  .101 &-.104 & .310 &.103 \\
    \midrule
    \multicolumn{5}{l}{PCC($\epsilon$, $\mu$)} \\
    \quad audio-eGeMAPS	&& .150 & .072 & .040 & -.024 \\ 
    \quad video-appearance	&& -.077 & .060 & -.077 & .059 \\ 
    \quad video-geometric	&&	.127 &.071 & .050 &.021 \\ 
    \bottomrule
    \end{tabular}
    \end{threeparttable}
  \end{table}

When training the networks jointly with input reconstruction (RE-based MTL) or perception uncertainty prediction (PU-based MTL), one can observe that the systems slightly outperform the baseline systems in nine out of twelve cases on the test set. 
This indicates that there is a substantial relationship between the two jointly learnt tasks. To be more specific, the representations from the last neural network hidden layer, which are learnt synchronously from the emotion prediction and other auxiliary tasks (\ie reconstructing the input or predicting the perception uncertainty), potentially further benefit the emotion prediction.

We further implemented the curriculum learning approach as well as its baseline by means of DNNs~\cite{Lotfian18-Curriculum} and GRU-RNNs. From Table~\ref{tab:beforePP}, 
it can be seen that the DNNs perform unsurprisingly worse than GRU-RNNs, mainly due to their limited capability of capturing the context information~\cite{Graves12-Supervised}. When feeding the data to the training model from a low-difficulty level to a high-difficulty level, the performance of the models is remarkably boosted in all scenarios. Nevertheless, it is still not competitive with the DDAT models in most cases. Moreover, it is observed that the GRU-RNN-based curriculum learning outperforms the DNN-based system mainly due to the learning capability of GRUs.

  The performance of the MTL systems is further enhanced by the proposed DDAT framework, as shown in Table~\ref{tab:afterPP}.
  In particular, the performance of the DDAT system for arousal and valence regressions respectively reaches CCC values of 0.694 and 0.422 with the audio-eGeMAPS feature set, 0.438 and 0.457 with the video-appearance feature set, and 0.400 and 0.501 with the video-geometric feature set. These results demonstrate that the DDAT systems significantly outperform ($p<.05$ via Fisher r-to-z transformation) the baseline method as well as the MTL approach (except in the case of valence regression with the audio-eGeMAPS feature set).  
  
Moreover, the systems using the proposed DDAT framework consistently outperform the curriculum learning approach, and they are competitive with, and in some cases even superior to, most other state-of-the-art methods, such as the strength modelling~\cite{Han17-Strength} and the `sparse coding (SC) + CNN + LSTM' systems (AVEC 2016 winner)~\cite{Brady16-Multi}. 
Despite the fact that the proposed systems are slightly worse than the end-to-end
system, which automatically extracts the representations from raw audio and video signals that retain complete pattern information, the DDAT framework can be incorporated with the end-to-end system in the future.

When comparing the two approaches used in the RE-based DDAT experiments, we find that adding the overall sum of the error [cf.~Fig.~\ref{fig:outline} (b)] leads to a better performance than adding the error vector [cf.~Fig.~\ref{fig:outline} (a)].
This is possibly attributable to the redundant dimensionality of the error vector, which meanwhile yields much noise in the network training. When comparing the RE-based DDAT and the PU-based DDAT, it is noticeable that the two approaches perform similarly. This suggests that both approaches achieve the same goal but in different ways. That is, both approaches successfully explore the difficulty information in the pattern learning process, whereas the RE-based and PU-based DDAT approaches measure the difficulty information by the data reconstruction-capability and by the data perception-uncertainty, respectively. Moreover, it is worth mentioning that the RE-based DDAT approach, in contrast to the PU-based DDAT, not only fits the subjective pattern recognition tasks (\eg emotion prediction in this work) but also holds the potential to be applied to objective tasks (\eg phoneme prediction). 

To investigate the contribution of the extracted difficulty information to the system performance improvement, we further calculated the correlation (in terms of PCC) between the values of the difficulty indicator (\ie the obtained RE or the PU) between the performance improvement. Specifically, the performance improvement $\Delta_c$ was computed as $\Delta_{c} = |\hat{y}_{bs} - {y}| - |\hat{y}_{DDAT} - {y}|$, given the target (gold standard) $y$ and the prediction of the DDAT system $\hat{y}_{DDAT}$ (or the baseline system $\hat{y}_{bs}$). 

The first three rows of Table~\ref{tab:relationship} show the obtained PCCs between the RE and the performance improvement [\ie PCC($\epsilon$, $\Delta_c$)]. These positive PCCs suggest that the difficulty information can help improve the model performance in the learning process. This conclusion confirms our previous findings in~\cite{Han17-Reconstruction}. Note that, in~\cite{Han17-Reconstruction}, the selected database has subjects that are different from the one in this article. 
Similar observations can be found when calculating the PCCs between the PU and the performance improvement [\ie PCC($\mu$, $\Delta_c$), as shown in the second three rows]. The PCCs are boosted to .384 and .440 in the development and test sets in the case of valence when using appearance-based visual features. 
In more detail, it can be seen that when using the RE-based DDAT approach, the achieved PCCs for arousal prediction are relatively higher than the ones for valence prediction in most cases. Nevertheless, an opposite observation is made when using the PU-based DDAT approach. This is probably due to the fact that arousal is more sensitive than valence to the expression strength or scale that potentially results in higher RE, whilst the valence is more associated with the subtle variations that easily mislead the judgement of annotators~\cite{Zeng09-Survey,Schuller13-Computational}. 

Furthermore, we calculated the PCCs between the obtained RE and PU, as shown in the last three rows in Table~\ref{tab:relationship}. Generally speaking, most of these PCCs are around zero, indicating the obtained RE is largely independent of the PU. This further implies that the proposed RE-based and PU-based DDAT strategies capture the different underlying phenomena. Thus, the combination of the two approaches is expected to deliver better performance. The related experiments and corresponding results are given in Section~\ref{subsec:dt}.

\subsection{Dynamic Tuning and Late Fusion}

\label{subsec:dt}
\begin{figure*}[!t]
\centering
    \subfigure[{arousal}]{
    \includegraphics[width=0.48\textwidth]{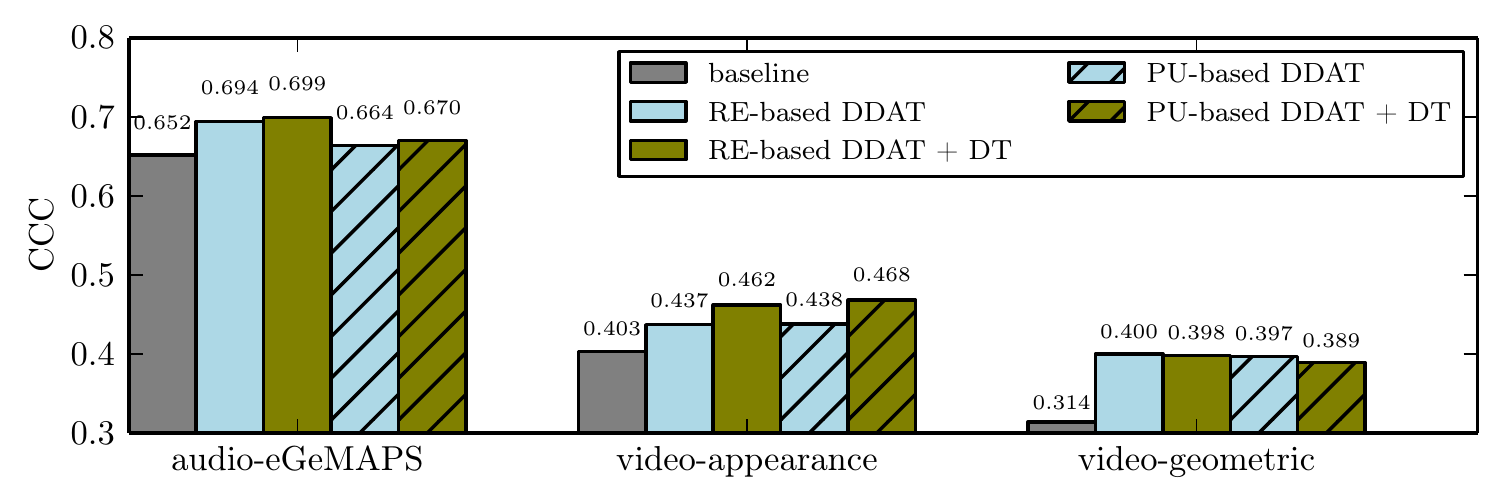}
    }
    \subfigure[{valence}]{
    \includegraphics[width=0.48\textwidth]{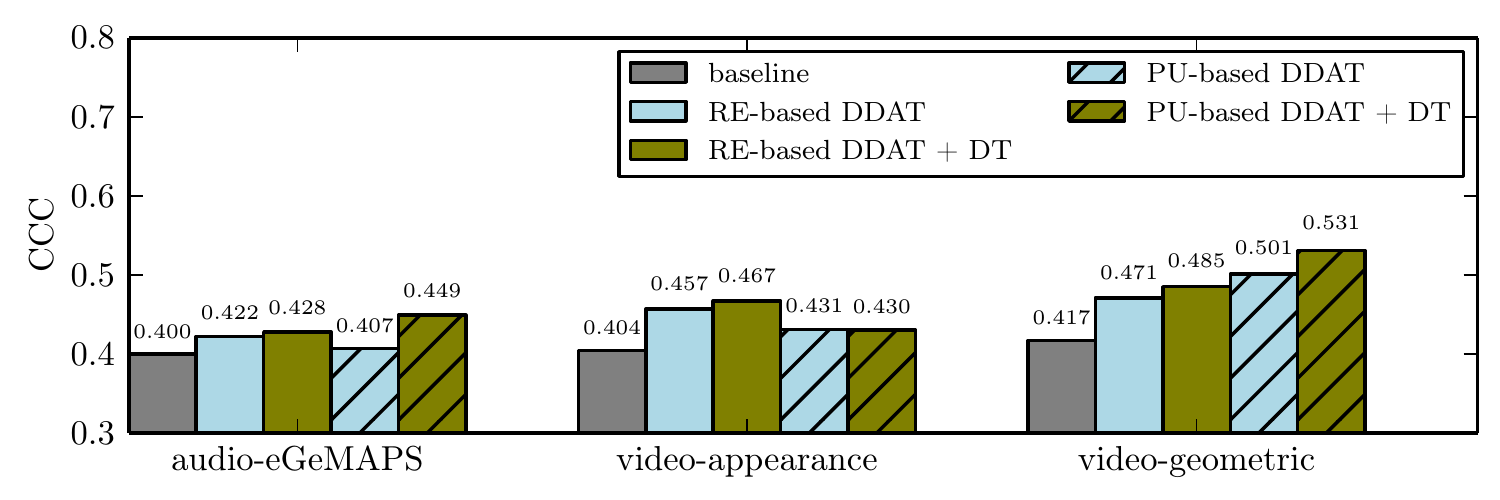}
    }
\caption{Performance comparison (CCC) between the single-task learning, the proposed dynamic difficulty awareness training approach based on reconstruction error (RE) or perception uncertainty (PU), and their dynamically-tuned (DT) versions. Results pertain to the test partition for both arousal (a) and valence (b) targets using three feature sets (audio-eGeMAPS, video-appearance, and video-geometric).}
\label{fig:dt}
\end{figure*}

Figure~\ref{fig:dt} illustrates the performances of the DDAT models with and without dynamic-tuning of the predictions.
Compared with the predictions without dynamic-tuning, the dynamically-tuned predictions yield gains in most cases. 
For instance, the best achieved CCC for arousal prediction increased from 0.684 to 0.699, using the RE-based DDAT system with the audio-eGeMAPS feature set, whereas for valence prediction it increased from 0.511 to 0.531, using the PU-based DDAT system with the video-geometric feature set.
The exceptions include the arousal predictions for both RE- and PU-based DDAT systems using the video-geometric feature set and the valence predictions for the PU-based DDAT system using the video-appearance feature set. In both cases, the differences remain minimal and insignificant via the aforementioned statistical test of Fisher z-to-r transformation. 

\begin{table}[!t]
\centering
\caption{Late fusion performance (CCC) in different fusion strategies (\ie modality-based, modality- and model-based, and dynamically-tuned modality- and model-based) for the \textit{dev}elopment and \textit{test} partitions of both {\em aro}usal and {\em val}ence regressions. The predictions are generated from the reconstruction-error-based DDAT framework ($P_{re}$) or the perception-uncertainty-based DDAT framework ($P_{pu}$); their dynamically-tuned versions ($P_{re, dt}$ or $P_{pu, dt}$); or the baseline model ($P_{bs}$). The best results achieved on the test set are in bold. Note that $P_{re}$, $P_{re, dt}$, $P_{pu}$, $P_{pu, dt}$, and $P_{bs}$ are the fused predictions from diverse information streams (\ie audio-eGeMAPS, video-appearance, and video-geometric). 
The 1st-3rd, 4th-5th, and 6th-8th result rows are respectively obtained from the modality-based, modality- and model-based, and dynamically-tuned modality- and model-based late fusion strategies.}

\label{tab:fusion}
  \begin{threeparttable}
  \begin{tabular}{p{.02\textwidth}p{.03\textwidth}p{.02\textwidth}p{.03\textwidth}p{.02\textwidth}cccc}
  \toprule
  \multicolumn{5}{c}{various late fusion approaches}        & \multicolumn{2}{c}{aro}   & \multicolumn{2}{c}{val}   \\
  $P_{re}$&$P_{re, dt}$&$P_{pu}$&$P_{pu, dt}$&$P_{bs}$        & dev   & test  & dev   & test          \\
  \midrule
  \multicolumn{8}{l}{\textit{modality-based}} \\
  &&&&\checkmark &.822 & .690 & .705 & .584 \\
  \checkmark&&&& & .853 & .763 & .738 & .615 \\
  &&\checkmark&& & .838 & .715 & .738 & .615 \\
  \midrule

  \multicolumn{8}{l}{\textit{modality- and model-based}} \\
  \checkmark&&\checkmark& &  & .860 & .761 & .755 & .639 \\
  \checkmark&&\checkmark&&\checkmark &    .864 & .752 & .766 & .653 \\
  \midrule

  \multicolumn{8}{l}{\textit{modality- and model-based (dynamically-tuned)}} \\
  &\checkmark&&& &   .853 & .761 & .739 & .621 \\
  &&&\checkmark&&   .819 & .721 & .733 & .631 \\
  &\checkmark&&\checkmark& &   .856 & \bf{.766} & .756 & .651 \\
  &\checkmark&&\checkmark&\checkmark &   .863 & .754 & .766 & \bf{.660} \\
  \midrule
  \multicolumn{9}{l}{\em{state of the art}} \\
  \multicolumn{5}{l}{\quad strength modelling~\cite{Han17-Strength} }                   & .808 & .685 & .671 & .554 \\
  \multicolumn{5}{l}{\quad end-to-end~\cite{Tzirakis17-End}} & .731 & .714 & .502 & .612 \\
    \multicolumn{9}{l}{\em{state of the art (+ physiology)}} \\
  \multicolumn{5}{l}{\quad feature selection + offset~\cite{He15-Multimodal} \tnote{a}}    &  .824& .747& .688& .609 \\
  \multicolumn{5}{l}{\quad SVR + offset~\cite{Valstar16-AVEC} \tnote{b}}   & .820 & .702 & .682 & .638 \\
  \multicolumn{5}{l}{\quad SC + CNN + LSTM~\cite{Brady16-Multi} \tnote{c}} & .862 & .770 & .750 & .687 \\

  \bottomrule
  \end{tabular}

  \begin{tablenotes}
  \scriptsize
  \item[a]  AVEC '15 winner
  \item[b]  AVEC '16 baseline
  \item[c]  AVEC '16 winner
  \end{tablenotes}

  \end{threeparttable}
\end{table}

\begin{figure*}[!t]
 \centering
  \includegraphics[width=0.95\textwidth, height=2in]{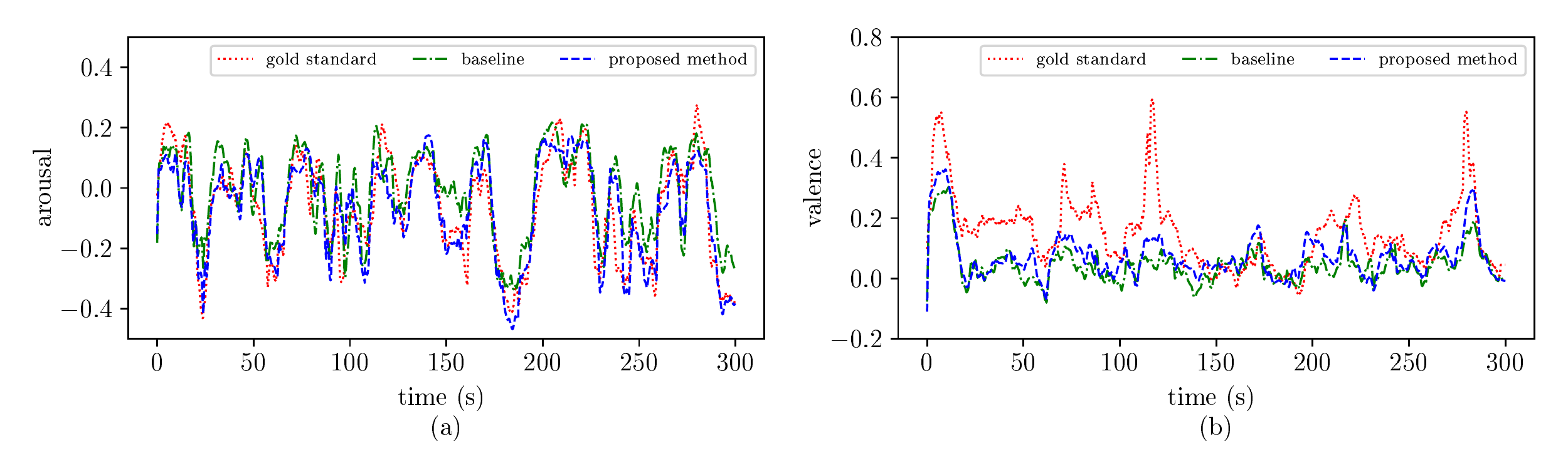}
 \caption{Automatic prediction of arousal (a) and valence (b) via audiovisual signals obtained with the best late fusion model for a random subject (\# 9) from the test partition.}
\label{fig:demo}
\end{figure*}

We then conducted a set of late fusions on the individual predictions produced by using different modalities and models. Table~\ref{tab:fusion} lists all scenarios (combinations) considered in our experiments as well the respective performance.
As can be seen in the table, the best performance on the test set for both arousal and valence was obtained when fusing the predictions from all \textit{modalities} and \textit{models}.
In this context, the best results on the test set have been achieved at 0.766 CCC for arousal and 0.660 CCC for valence.
These results beat most of the latest reported results from the same data, and they are close to the best result presented in AVEC 2016~\cite{Brady16-Multi} (\ie 0.770 and 0.687 of CCCs for arousal and valence prediction), despite this system also utilising an additional modality (physiological features).
An illustration of the performance of the best DDAT system compared to the baseline system and the gold standard is depicted in Fig.~\ref{fig:demo} (data from a random subject from the test partition).
Generally, it can be seen that our predictions are closer to the gold standard, especially in the region that has relative peak values.

\begin{figure}[!t]
    \centering
    \subfigure[{information stream}]{
    \includegraphics[width=0.22\textwidth]{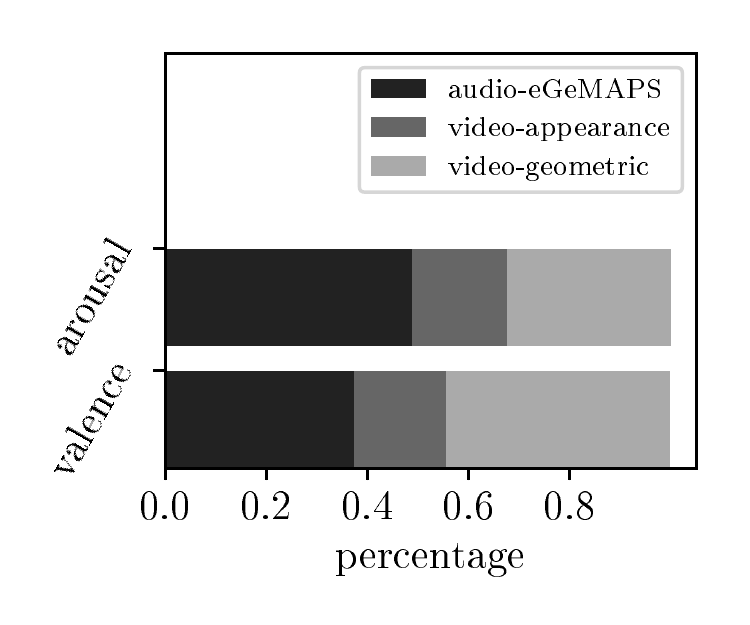}
    }
    \subfigure[{model}]{
    \includegraphics[width=0.22\textwidth]{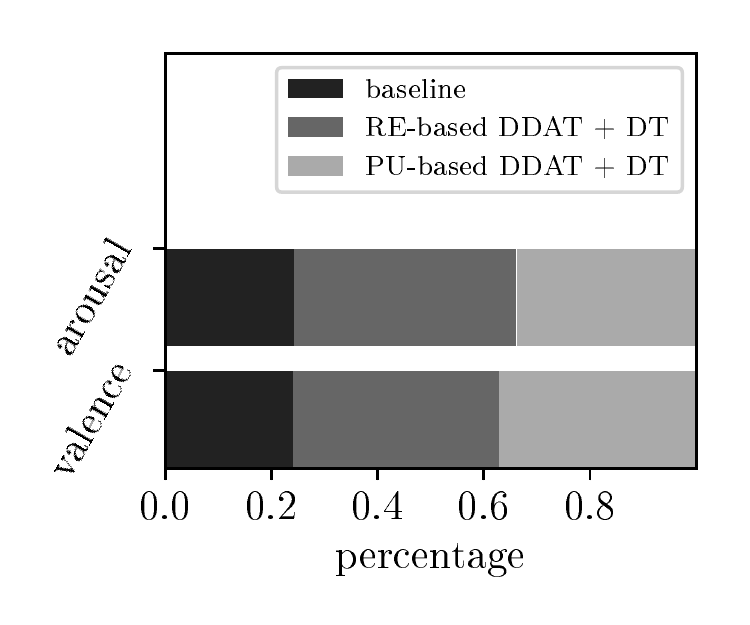}
    }
    \caption{Percentage of the contribution of each information stream (a) or model (b) for achieving the best arousal or valence predictions.}
    \label{fig:percentage}
\end{figure}

In order to analyse the importance of each modality and model, we calculated their contributions to the arousal and valence predictions of the respective best performing models.
Fig.~\ref{fig:percentage} depicts their contributions. 
For arousal prediction, the acoustic features play a more important role than the visual features, whereas the opposite happens for valence prediction.
It is also expected that the RE-based and PU-based DDAT systems contribute more than the baseline systems to the final predictions.
Furthermore, the PU-based DDAT system is slightly more important for valence prediction than it is for arousal prediction.
This might be due to the fact that prediction of emotional valence is much more difficult than arousal for audio modality~\cite{Gunes10-Automatic,Zhang15-Cooperative,Coutinho18-Shared}. 

\section{Conclusions and Future Work}
\label{sec:conclusion}
In contrast to previous studies that aimed to explore the `strength' or overcome the `weakness' of modelling, we for the first time investigated exploiting the difficulty (weakness) information straightforwardly in the learning process for continuous emotion prediction. To extract the difficulty information, we proposed two strategies based on either the ontology of modelling or the content to be modelled. The two types of information separately measure the learning difficulty of a model by reconstructing its input, or the `hardness' of the data to be learnt by predicting their perception uncertainty. This information indicated by an index was then concatenated into the original features to update the inputs.
The proposed methods were systematically evaluated on a benchmark database RECOLA~\cite{Valstar16-AVEC}. Experimental results have demonstrated that the proposed methods clearly improve the prediction performance of a model by evolving the difficulty information into its learning process.  

Going beyond the traditional curriculum learning and boosting approaches that are specifically designed for discrete pattern recognition tasks, the proposed Dynamic Difficulty Awareness Training (DDAT) approaches can particularly learn well the sequential pattern, such as the continuous emotion prediction in this article. When involving either the input reconstruction error information or the emotion perception uncertainty information, we find that the neural networks can better perform. Nevertheless, it is worth noting that the perception uncertainty is merely defined for a subjective pattern recognition task. For an objective task, it might be reasonable to alternatively employ the prediction uncertainty.

In future, we will continue investigating the efficiency of the proposed DDAT in discrete pattern predictions. Additionally, we will investigate the approaches for which the difficulty information could be possibly used as the prediction weights. Moreover, end-to-end structures that are designed to automatically extract representations have attracted increasing attention, and they are starting to show promising performance. Therefore, an advanced end-to-end framework will be considered in our system as well. In more detail, with respect to the perception-uncertainty-based DDAT end-to-end system, we can simply replace the GRN-RNNs with an end-to-end network while all other inputs and outputs remain. With respect to the reconstruction-error-based end-to-end system, we may consider reconstructing the high-level representations rather than the raw signals when extracting the reconstruction error information (\ie the difficulty indicator).

\section*{Acknowledgment}

This work was supported by a TransAtlantic Platform ``Digging into Data'' collaboration grant (ACLEW: Analysing Child Language Experiences Around The World), with the support of the UK's Economic \& Social Research Council through the research Grant No.~HJ-253479 (ACLEW), 
and the European Union's Horizon 2020 Programme through Research \& Innovation Action No.~645094 (SEWA) and No.~645378 (ARIA-VALUSPA).

\bibliographystyle{IEEEtran}
\bibliography{ddaref}

\begin{IEEEbiography}[{\includegraphics[width=1in,height=1.25in,clip,keepaspectratio]{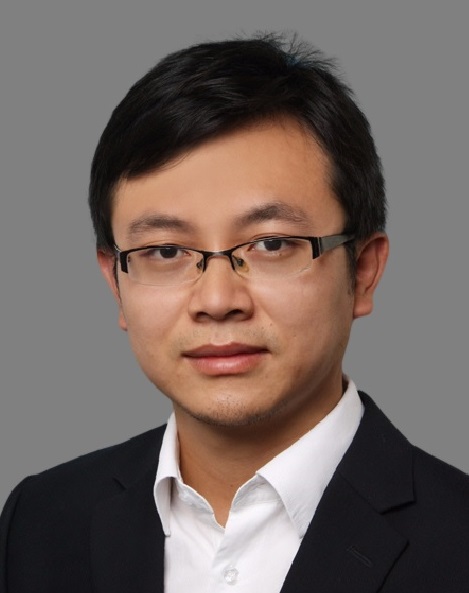}}]
{\bf Zixing Zhang} (M'15)
received his master degree in physical electronics from the Beijing University of Posts and Telecommunications (BUPT), China, in 2010, and his PhD degree in computer engineering from Technical University of Munich (TUM), Germany, in 2015. Currently, he is a research associate with the Department of Computing at the Imperial College London (ICL), UK, since 2017. Before that, he was a postdoctoral researcher at the University of Passau, Germany, from 2015 to 2017. He has authored about seventy publications in peer-reviewed books, journals, and conference proceedings to date, and has organised special sessions, such as at the IEEE 7th Affective Computing and Intelligent Interaction (ACII) in 2017 and at the 43nd IEEE International Conference on Acoustics, Speech, and Signal Processing (ICASSP) in 2018. Dr Zhang serves as a reviewer for leading-in-their fields journals such as  IEEE T-NNLS, IEEE T-CYB, IEEE T-AC, IEEE T-MM, IEEE T-ASLP, Speech Communication, and Computer Speech \& Language. His research interests lie in deep learning, weekly supervised learning, and transfer learning for intelligent and robust speech analysis, such as emotion recognition. 
\end{IEEEbiography}

%
\begin{IEEEbiography}[{\includegraphics[width=1in,height=1.25in,clip,keepaspectratio]{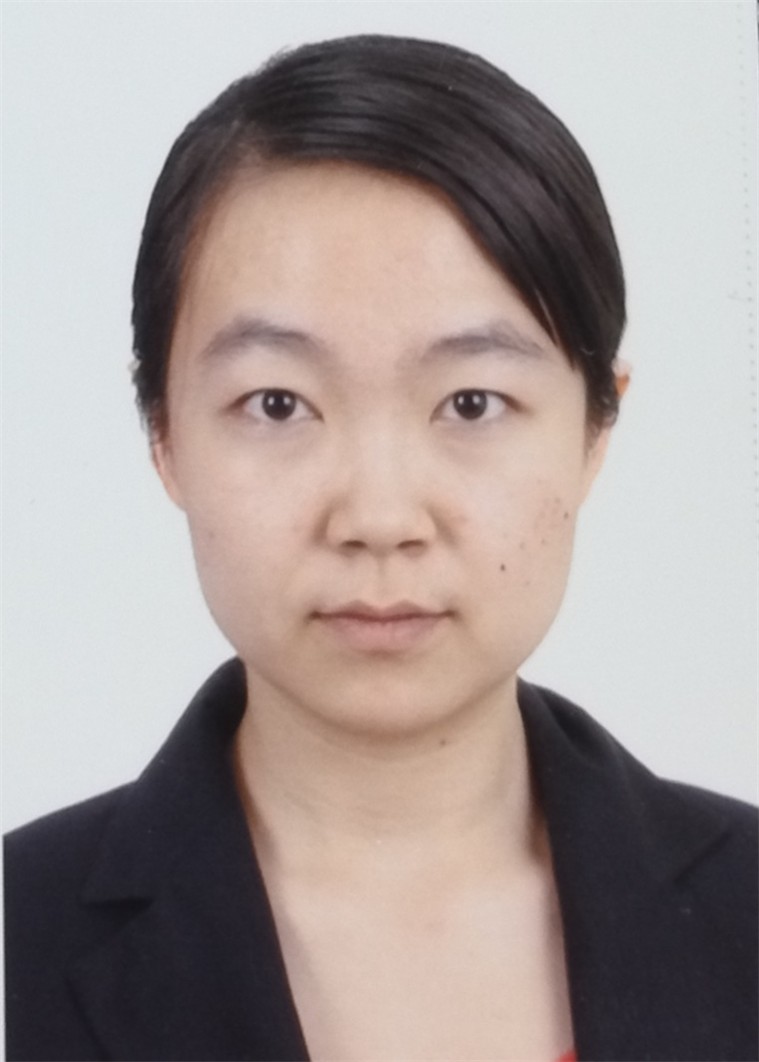}}]
{\bf Jing Han} (S'16) received her bachelor degree (2011) in electronic and information engineering from Harbin Engineering University (HEU), China, and her master degree (2014) from Nanyang Technological University, Singapore. She is now working as a doctoral student with the ZD.B Chair of Embedded Intelligence for Health Care and Wellbeing at the University of Augsburg, Germany,
involved in the EU's Horizon 2020 programme SEWA. She reviews regularly for IEEE Transactions on Cybernetics and IEEE Signal Processing Letters. Her research interests are related to deep learning for multimodal affective computing and  health care.
\end{IEEEbiography}

\begin{IEEEbiography}
[{\includegraphics[width=1in,height=1.25in,clip,keepaspectratio]{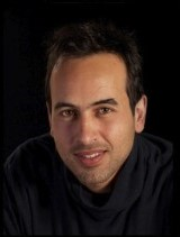}}]
{\bf Eduardo Coutinho} graduated from the University of Porto (Portugal) in 2003, and completed his Ph.D in Computer and Affective Sciences in 2009. Since then, Coutinho has worked in the interdisciplinary fields of music psychology and affective sciences at the University of Sheffield (UK), the Swiss Center for Affective Sciences (Switzerland), the Technical University of Munich (Germany) and Imperial College London (UK). Currently, Coutinho is a Lecturer in Music Psychology at the University of Liverpool (UK) and a Research Associate in the ZD.B Chair of Embedded Intelligence for Health Care and Wellbeing at the University of Augsburg (Germany). In his research, Coutinho focuses on the study of emotional experiences with music and the links between the communication of emotion by music and the tone of voice. Currently, he is also engaged in the development of methods and tools that permit the use music for improving different aspects of well-being in everyday life. In 2013, he received the Knowledge Transfer Award from the Swiss National Center of Competence in Research in Affective Sciences, and in 2014 he was the recipient of the Young Investigator Award from the International Neural Network Society.
\end{IEEEbiography}

\begin{IEEEbiography}[{\includegraphics[width=1in,height=1.25in,clip,keepaspectratio]{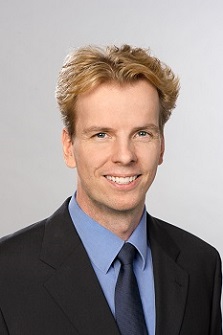}}] 
{\bf Bj\"orn~Schuller} (M'05-SM'15-F'18) received his diploma in 1999, his doctoral degree for his study on Automatic Speech and Emotion Recognition in 2006, and his habilitation and Adjunct Teaching Professorship in the subject area of Signal Processing and Machine Intelligence in 2012, all in electrical engineering and information technology from TUM in Munich/Germany. {He is Professor in Machine Learning in the Department of Computing at the Imperial College London/UK, where he heads GLAM -- the Group on Language, Audio \& Music, Full Professor and head of the ZD.B Chair of Embedded Intelligence for Health Care and Wellbeing at the University of Augsburg/Germany, and CEO of audEERING.  
He was previously full professor and head of the Chair of Complex and Intelligent Systems at the University of Passau/Germany. }
Professor Schuller is President-emeritus of the Association for the Advancement of Affective Computing (AAAC), elected member of the IEEE Speech and Language Processing Technical Committee, and Fellow of the IEEE, and Senior Member of the ACM. He (co-)authored 5 books and more than 700 publications in peer-reviewed books, journals, and conference proceedings leading to more than overall 19\,000 citations (h-index = 66). Schuller is co-Program Chair of Interspeech 2019, repeated Area Chair of ICASSP, and Editor in Chief of the IEEE Transactions on Affective Computing next to a multitude of further Associate and Guest Editor roles and functions in Technical and Organisational Committees.\end{IEEEbiography}

\end{document}